\documentclass{article}

\usepackage{arxiv}

\usepackage[utf8]{inputenc} 
\usepackage[T1]{fontenc}    
\usepackage{hyperref}       
\usepackage{url}            
\usepackage{booktabs}       
\usepackage{amsfonts}       
\usepackage{nicefrac}       
\usepackage{microtype}      
\usepackage{lipsum}		
\usepackage{graphicx}
\usepackage{natbib}
\usepackage{doi}
\usepackage[capitalise]{cleveref}
\usepackage{subfig}

\title{xVLM2Vec: Adapting LVLM-based embedding models to multilinguality using Self-Knowledge Distillation}

\date{} 					

\author{Elio Musacchio \\
	Department of Computer Science\\
	University of Bari Aldo Moro\\
	Bari, Italy \\
	\texttt{elio.musacchio@uniba.it} \\
	\And{Lucia Siciliani} \\
	Department of Computer Science\\
	University of Bari Aldo Moro\\
	Bari, Italy \\
	\texttt{lucia.siciliani@uniba.it} \\
	\And{Pierpaolo Basile} \\
	Department of Computer Science\\
	University of Bari Aldo Moro\\
	Bari, Italy \\
	\texttt{pierpaolo.basile@uniba.it} \\
	\And{Giovanni Semeraro} \\
	Department of Computer Science\\
	University of Bari Aldo Moro\\
	Bari, Italy \\
	\texttt{giovanni.semeraro@uniba.it} \\
}

\renewcommand{\headeright}{}
\renewcommand{\undertitle}{}
\renewcommand{\shorttitle}{}

\hypersetup{
pdftitle={xVLM2Vec: Adapting LVLM-based embedding models to multilinguality using Self-Knowledge Distillation},
pdfauthor={Elio Musacchio},
pdfkeywords={LLM, LVLM, Multimodal},
}

\begin{document}
\maketitle

\begin{abstract}
	In the current literature, most embedding models are based on the encoder-only transformer architecture to extract a dense and meaningful representation of the given input, which can be a text, an image, and more. With the recent advances in language modeling thanks to the introduction of Large Language Models, the possibility of extracting embeddings from these large and extensively trained models has been explored. However, current studies focus on textual embeddings in English, which is also the main language on which these models have been trained. Furthermore, there are very few models that consider multimodal and multilingual input. In light of this, we propose an adaptation methodology for Large Vision-Language Models trained on English language data to improve their performance in extracting multilingual and multimodal embeddings. Finally, we design and introduce a benchmark to evaluate the effectiveness of multilingual and multimodal embedding models.
\end{abstract}

\keywords{LLM \and LVLM \and Multimodal}

\section{Introduction}
\label{sec:intro}
After the introduction of the Transformer architecture by \citet{vaswani2017attention}, researchers have started focusing on methods to leverage the latent representations learned by these models.
One of the most popular models in the textual domain is surely \textsc{BERT} \citep{devlin-etal-2019-bert}.
Thanks to its training strategy (exploiting bi-directional attention mask, masked token prediction and next sentence prediction), the model can learn semantically meaningful deep representations of input texts.
After the release of BERT, vision models, which previously focused on \textit{Convolutional Neural Networks} (e.g. \textsc{ResNet} \citep{he2016deep}), also started employing transformer-based architectures.
The most important example in the literature is the Vision Transformer (\textsc{ViT}) \citep{dosovitskiy2021imageworth16x16words}, which implemented the same logic as \textsc{BERT}, but, instead of splitting text into tokens, it splits images into fixed-size patches that are then used as input.
Nevertheless, these models only focused on encoding a single modality, textual for \textsc{BERT} and visual for \textsc{ViT}. Therefore, comparing embeddings across the two spaces learned by these models is impossible since they are substantially different.
The \textsc{CLIP} \citep{radford2021learning} model was developed to overcome this issue.
Using contrastive-learning, it learns a common embedding space where the same concepts from different modalities have a similar dense representation. 

After the release of \textsc{ChatGPT} \citep{openai2024gpt4technicalreport} and the massive research interest in \textit{Large Language Models} (\textsc{LLM}s), some works started focusing on converting decoder-only transformer models to embedding models.
However, since they are trained on the auto-regressive objective using a causal attention mask, they are not fit for embedding extraction right away. \textsc{LLM2Vec} \citep{llm2vec} and \textsc{E5} \citep{wang2022text} both introduced training strategies for this purpose.
Despite this, there are still gaps in the current literature. Specifically, \textsc{LLM}-based encoder models only consider the textual modality. While there are \textit{Large Vision-Language Models} (\textsc{LVLM}s) that extend decoder-only models to understand images (e.g. \textsc{LLaVA} \citep{liu2024visual}), there are very few strategies which leverage them for embedding extraction. Moreover, there is no multilingual and multimodal \textsc{LVLM} embedding model trained using a distillation approach.

The contributions of this work are the following:
\begin{itemize}
    \item We introduce \textsc{xVLM2Vec}, an adaptation methodology using a Self-Knowledge Distillation approach to improve the quality of multilingual embeddings learned by an English only LVLM embedding model while also preserving the quality of the English embeddings;
    \item We introduce \textsc{MMMEB} (Massive Multimodal and Multilingual Embedding Benchmark), the first benchmark to evaluate the quality of multimodal and multilingual embedding models;
    \item We release all the resources obtained from our experiments (data, models and code) to boost the research in multimodal and multilingual embedding models.
\end{itemize}

\section{Related Works}
\label{sec:related_works}

In the current literature, there are many embedding models leveraging the innovations that have been brought by \textsc{LLM}s.
For instance, \textsc{LLM2Vec} \citep{llm2vec} is a methodology to train decoder-only transformer models for embedding extraction.
In particular, the causal attention mask is replaced by a bi-directional one and two training stages are performed (masked next token prediction and unsupervised contrastive learning).
Another example is represented by \textsc{E5} \citep{jiang2024e5}, which performs a contrastive pre-training stage followed by fine-tuning, extracting embeddings using average pooling of the output layer.

Furthermore, extending embedding models to multilinguality has always been of research interest; some examples in the textual domain include \textsc{mBERT} \citep{devlin-etal-2019-bert} and \textsc{XLM-Roberta} \citep{rob}.
More recently, \textsc{mE5} \citep{wang2024multilingual} has been released, which uses the same training approach as the \textsc{E5} models on a multilingual data mixture.
However, these models tend to produce embedding representations that are not comparable across languages due to the training performed on the same objective as English models with a training mixture consisting of diverse languages.
For example, \textsc{XLM-Roberta} was trained on 2.5TB of CommonCrawl data in 100 languages. However, embeddings for the same concept in different languages may not have the same dense representation since there is no constraint during training to align them.
In light of this, \citet{reimers-gurevych-2020-making} proposed a \textit{Knowledge Distillation} approach to make monolingual embedding models multilingual.
The idea behind Knowledge Distillation is to perform transfer-learning by training a smaller model (called \textit{student}) using the outputs of a larger pre-trained model (called \textit{teacher}).
By doing so, the knowledge of the larger model is compressed into the smaller one.
\citet{reimers-gurevych-2020-making} exploited this approach by training the student model to optimize the embedding representation of an English input and its translation to multiple languages so that they would be as close as possible to the English representation learned by the teacher model.
This approach was then further exploited in later works, like \textsc{NLLB-LLM2Vec} \citep{schmidt2024selfdistillationmodelstackingunlocks}, where the encoder of the \textsc{NLLB} model (an encoder-decoder machine translation model) is merged into the \textsc{LLM}, using the representations learned by the \textsc{NLLB} for all of its languages as input to the \textsc{LLM}.
Training is performed in two stages: \textit{Self-Supervised General Adaptation} to merge the \textsc{NLLB} encoder using a self-supervised approach where the original \textsc{LLM} acts as teacher, and \textit{Task-Specific Distillation}, still performing self-supervision but using a fine-tuned version of the original \textsc{LLM}.

In the vision-language domain, instead, embedding models mainly focused on \textsc{CLIP} \citep{radford2021learning} and \textsc{SigLIP} \citep{zhai2023sigmoid} since they are both able to learn a common embedding space for both visual and textual inputs. The former used a contrastive loss for training, while the latter used a pairwise sigmoid loss. 
Many works have performed multilingual adaptation of the \textsc{CLIP} model. 
For example, \citet{carlsson-EtAl:2022:LREC} used a Knowledge Distillation approach to make the text encoder of \textsc{CLIP} multilingual.
The goal is to keep intact the multimodal embedding space learned by \textsc{CLIP}, this is obtained by training the text encoder to support multilingual inputs.
\textsc{E5-V} \citep{jiang2024e5} has also been released, where multimodal representations are unified using a prompt-based approach (removing the modality encoder and projector in the original \textsc{LVLM}).
Recently, \textsc{VLM2Vec} \citep{jiang2024vlm2vec} was released, a contrastive training framework to convert any \textsc{LVLM} into an embedding model. 
The authors released checkpoints for LoRA and full parameter fine-tuning of the \textsc{Phi-3.5-vision-instruct} \citep{abdin2024phi3technicalreporthighly} model. 
The main advantage of embeddings extracted using this approach w.r.t. \textsc{CLIP} or \textsc{SigLIP} models is the degree of generalization.
Indeed, since \textsc{Phi} is an instruct model, it is possible to use instructions to encode task information within the embedding. 
Finally, some decoder-only multilingual instruct models have been released, like \textsc{mBlip} \citep{geigle-etal-2024-mblip}, a \textsc{BLIP-2} model trained on a machine-translated dataset covering 96 languages, and \textsc{Pangea} \citep{yue2024pangeafullyopenmultilingual}, a \textsc{LVLM} trained on a curated training mixture containing cultural aspects for 39 languages.

Despite the research interest in embedding extraction from \textsc{LLM}s and multilingual embeddings, no work has studied the effect of multilingual adaptation and training for embedding extraction on a \textsc{LVLM}. 

\section{Training Methodology}
\label{sec:train_method}

\begin{figure*}[ht]
  \centering
   \includegraphics[width=\textwidth]{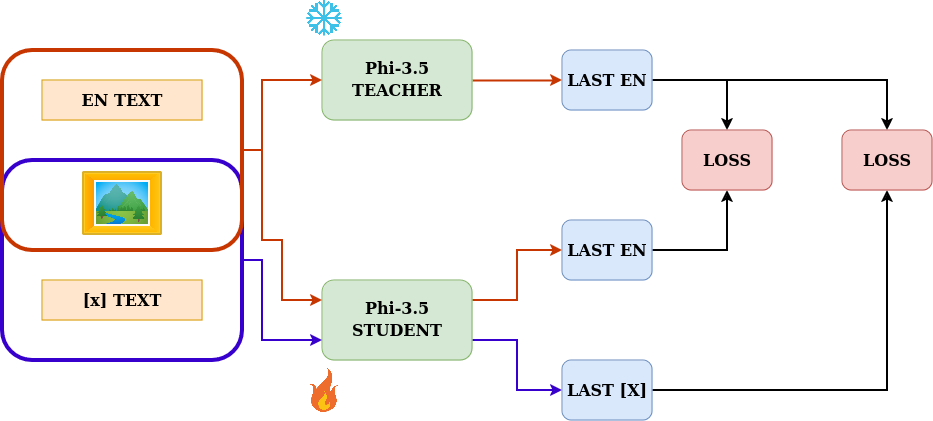}
   \caption{Overview of the proposed Self-Knowledge distillation approach. $x$ refers to a non-English language out of the ones we consider in our training mixture (\textit{French}, \textit{German}, \textit{Italian} and \textit{Spanish}). Training is done on a parallel dataset of pairs of English language and $x$ language texts. The teacher model is kept frozen while the student model's weights are updated.}
   \label{fig:distillation}
\end{figure*}

For our adaptation methodology, we apply a \textit{Self-Knowledge Distillation} approach, inspired by the \textit{Knowledge Distillation} approach introduced in the textual domain by \citet{reimers-gurevych-2020-making}, where the teacher $T$ and the student $S$ are both copies of the same model.
Both $T$ and $S$ are embedding models that output a dense representation for a given input, which is a vector in $R^{m\times d}$ where $m$ is the sequence length of the input and $d$ is the embedding dimension of the model.
Specifically, given two text inputs $x, y$, where $x$ is a text written in the English language and $y$ is the translation of $x$ to a non-English language, the loss is computed as:

\begin{equation}
  loss_e(x, y) = loss_d(pool(T(x)), pool(S(x))) + loss_d(pool(T(x)), pool(S(y)))
  \label{eq:loss_last}
\end{equation}

\begin{equation}
    loss(x, y) = \frac{loss_e(x, y)}{2}
    \label{eq:loss_weighted}
\end{equation}

\noindent where $pool\colon R^{m\times d} \mapsto R^{1\times d}$ is a function used to map the embeddings to a single dense vector, $loss_d$ is a distance-based loss used to quantify the differences between the extracted embeddings.
The idea is to use $loss_e$ for two purposes: 1) to keep as faithful as possible the English embeddings of $S$ to the ones learned by $T$; 2) to make non-English embeddings of $S$ as close as possible to the English ones learned by $T$. The first and second addendum have these objectives, respectively. We then compute $loss$ as the average of the two addendums.

For $loss_d$, we use the \textbf{Mean Squared Error}, that is

\begin{equation}
  loss_d(x, y) = \frac{1}{d} \sum_{i=i}^{d}{(x_i - y_i)^2}
  \label{eq:loss}
\end{equation}

We apply this approach using \textsc{VLM2Vec} \citep{jiang2024vlm2vec}. Specifically, we use the LoRA checkpoint trained with a batch size of 1024 since it is a good compromise between performance and model size.
The embeddings for the original model are extracted from the last token in the sequence, therefore we do the same during training.
We keep the teacher's weights frozen while updating the ones of the student model. We use a full-parameter training approach by leveraging \textit{Fully Sharded Data Parallel} (\textsc{FSDP}) \citep{zhao2023pytorch}.
We train for 1 epoch with an effective batch size of 128 and a learning rate of 1e-5. Training is performed on a compute node consisting of four A100 64GB VRAM GPUs. An overview of the training approach is illustrated in \cref{fig:distillation}.

For the objectives of \cref{eq:loss_weighted} to work, it is imperative to have a parallel corpus. That is, the non-English input must directly translate the English one. We want to leverage the same dataset used to train the original model so that we are sure that the concepts learned initially by the English model will also be learned by its multilingual version. However, since it is only in English, we decided to rely on machine-translation.

\section{Translation Methodology}
\label{sec:translation}

\begin{figure*}[ht]
  \centering
   \includegraphics[width=\textwidth]{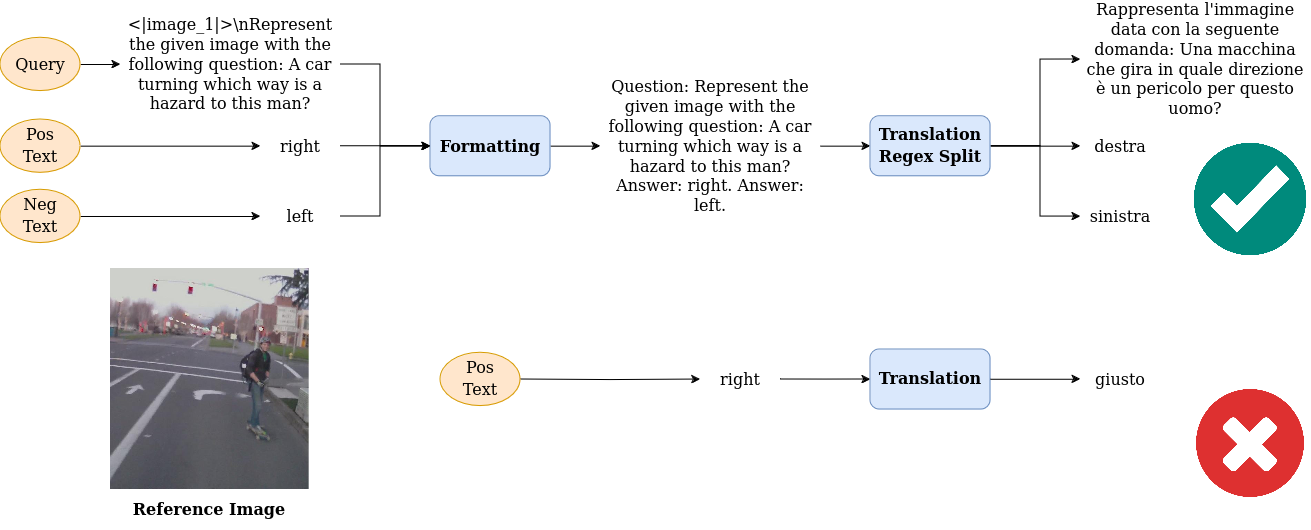}
   \caption{Comparison of two possible translation approaches considering a translation from English to Italian. In the example above, the translation methodology described in  \cref{sec:translation} is applied, while a direct translation methodology is applied in the example below. The term "right", which in this case refers to the concept of "direction" (e.g. "The car is on the right side"), is correctly translated in the first example, while it is not correctly translated in the second, where instead the translation model translates "right" with the concept of "correctness" (e.g. "Yes, that's right!"). These examples are obtained from the translation model \textsc{MADLAD-400 3B}. The example is from an actual instance in the training dataset from the \textsc{A-OKVQA} task.}
   \label{fig:translation}
\end{figure*}

In the original work by \citet{jiang2024vlm2vec}, the authors introduced \textsc{MMEB} (Massive Multimodal Embedding Benchmark), a benchmark consisting of train and test splits covering a wide array of tasks for multimodal embedding models.
Specifically, the benchmark focuses on four meta-tasks: \textit{Classification}, \textit{VQA}, \textit{Retrieval} and \textit{Visual Grounding}. 

We first select a subset of tasks to translate the \textsc{MMEB} train set.
We need to make this selection since some tasks in the original dataset are inappropriate for translation.
Specifically, tasks in which the images contain English text (e.g. \textsc{DocVQA} contains VQA instances which require understanding of document text in an image) and focus on text extraction or comprehension would completely lose meaning when translated into a different language.
Additionally, we also remove tasks where the text formatting makes translation far too complex (e.g. \textsc{VisDial} instances are formatted by adding "Q:" and "A:" string pairs in the text).
Specifically, we remove the following tasks: \textsc{ChartQA}, \textsc{DocQA}, \textsc{HatefulMemes}, \textsc{InfographicsVQA}, \textsc{ScienceQA} and \textsc{VisDial}.
Finally, we have 20 tasks out of the 26 original ones after filtering.
Note that we are leveraging a version of \textsc{MMEB} where 26 tasks are available, differently from the default one where 20 are available. Specifically, the original dataset is available on Huggingface\footnote{\url{https://huggingface.co/datasets/TIGER-Lab/MMEB-train/tree/0c3f4b828d347c4e8508339f99530f6c820061fd}}.
Then, we select a subset of non-English languages to consider. We focus on \textit{French}, \textit{German}, \textit{Italian} and \textit{Spanish}, which are all popular European languages. Finally, we want to have a training mixture that is as task-balanced as possible so that the student model can align well on all tasks.
To do so, we select the first 10,000 instances of each task out of the 20 ones we selected.
We fix this threshold since there are tasks that have less than 10,000 instances.
This ensures that these tasks are still represented while also covering a wide array of examples from the others.
We rely on one of the most recent machine-translation models, that is \textsc{MADLAD-400 3B} \citep{kudugunta2024madlad}, for automatic translation.
However, the machine-translation model only uses the textual modality as input, therefore, directly translating without the context given by the image may result in translation errors. To reduce the possibility of translation errors, we concatenate the query, the positive target text and the negative target text (the latter only when available). For each instance of the dataset, we append a "Question:" string before the query contents and an "Answer:" string before each of the target texts (both positive and negative). The intuition is that the translation of the query and target texts together will be more precise, thanks to the additional textual content. During the translation process, we replace, when present, the "\textless$\vert$image\_1$\vert$\textgreater\textbackslash n" string, which is used by \textsc{VLM2Vec} as a placeholder for the image embeddings. After translation, we extract query and target texts using a regex-based approach, to separate the translated output from the "Question:" and "Answer:" strings. The instance is discarded when, due to translation errors, it is impossible to extract the translated strings correctly. An example of this translation process is provided in \cref{fig:translation}, with an example of direct translation of the same instance. While this methodology is still imperfect, it adds additional context to the translation process, reducing the possibility of translation errors while exploiting a text-only machine-translation model. After collecting the output of the machine-translation process, we keep only the query and positive target text (discarding the negative target text) as independent instances. The complete parallel corpus consists of a total of 1,295,651 instances.


For evaluation, there is no benchmark like \textsc{MMEB} for multilingual and multimodal models covering the same meta-tasks. We could perform the same procedure applied for our train set, using a machine-translation model to translate the test sets of \textsc{MMEB}. However, the evaluation would be unfair w.r.t. other models since any machine-translation error learned by the model may unfairly boost its performance during evaluation if the test set also contains these errors. Therefore, we introduce the \textit{Massive Multimodal and Multilingual Embedding Benchmark} (\textsc{MMMEB}), the first benchmark for multimodal and multilingual embedding models.

\section{Benchmark Creation}
\label{sec:benchmark}

\begin{table*}[htb]
  \centering
  \begin{tabular}{@{}llll@{}}
    \toprule
    \textbf{Dataset} & \textbf{Cardinality} & \textbf{Supported Languages} & \textbf{Supported Tasks} \\
    \midrule
    \textsc{Crossmodal-3600} & 3600 & \textsc{ALL} & \textsc{T2I}, \textsc{I2T} \\
    \textsc{XTD-10} & 1000 & \textsc{ALL} & \textsc{T2I}, \textsc{I2T} \\
    \textsc{MaXM} & 257 & \textsc{EN} & \textsc{VQA} \\
    \textsc{MaXM} & 264 & \textsc{FR} & \textsc{VQA} \\
    \textsc{Flickr30k Entities} & 4042 & \textsc{EN} & \textsc{VG} \\
    \textsc{Flickr30k Entities} & 2825 & \textsc{FR} & \textsc{VG} \\
    \textsc{Imagenet-1k} & 1000 & \textsc{ALL} & \textsc{C} \\
    \bottomrule
  \end{tabular}
  \caption{List of datasets used alongside supported tasks and languages. When \textsc{ALL} is used, it means that the dataset supports all languages considered in the benchmark (\textsc{DE}, \textsc{FR}, \textsc{EN}, \textsc{ES}, \textsc{IT}).}
  \label{tab:dataset}
\end{table*}

To create \textsc{MMMEB}, we decided to rely exclusively on multilingual datasets that have been either hand-written by humans or manually checked to reduce the possibility of translation errors in the evaluation datasets as much as possible. We select the following datasets: 

\begin{itemize}
    \item \textbf{\textsc{Crossmodal-3600}} \citep{ThapliyalCrossmodal2022}: a geographically-diverse set of 3,600 images annotated by humans. Since each image has multiple captions, we always select the first one for each language.
    \item \textbf{\textsc{XTD10}} \citep{aggarwal2020towards, rajendran-etal-2016-bridge}: dataset consisting of 1,000 multilingual image captions translated from MSCOCO2014 through a crowdsourcing platform.
    \item \textbf{\textsc{MaXM}} \citep{changpinyo2023maxm}: multilingual visual question answering dataset created by exploiting the captions in \textsc{Crossmodal-3600}.
    \item \textbf{\textsc{Flickr30k entities}} \citep{flickrentitiesijcv}: dataset consisting of captions annotated with bounding boxes of the objects they reference within the image. We crop the images by using the bounding boxes in the test split and perform the same process for its French translation \citep{flickr30k_french}.
    \item \textbf{\textsc{Imagenet-1k}} \citep{imagenet15russakovsky}: image dataset organized according to the WordNet hierarchy. We retrieve the translated class mappings from \textsc{Babel-ImageNet} \citep{geigle2023babelimagenet}, then we select the first 1,000 instances from the validation split (since the original test split does not provide the labels) and apply the translated mapping.
\end{itemize}

After collecting all datasets, we filter them to include only the languages considered in our train set, that are \textit{English}, \textit{French}, \textit{German}, \textit{Italian} and \textit{Spanish}.
Not all datasets support all of these five languages; when this happens, we only consider the subset of available languages. After filtering the test splits, we must format them for vision-language tasks. We decide to focus on the following tasks:
\begin{itemize}
    \item \textbf{Image to Text Retrieval} (\textsc{I2T}): given an image and a collection of candidate texts, the model needs to select the candidates related to the given image;
    \item \textbf{Text to Image Retrieval} (\textsc{T2I}): given a text and a collection of candidate images, the model needs to select the candidates related to the given text;
    \item \textbf{Visual Question Answering} (\textsc{VQA}): given an image, a question related to the image and a collection of possible answers, the model needs to select the candidates which correctly answer the given question;
    \item \textbf{Visual Grounding} (\textsc{VG}): given an image and an object in the image (represented by a textual string), the model needs to select the crop of the given image that contains the given object;
    \item \textbf{Classification} (\textsc{C}): given an image and a set of possible classes, the model needs to select the correct class for the image.
\end{itemize}

For each task and dataset, we must create a pool of candidate items, which can be either texts or images, depending on the task. For each instance $i$ in the dataset $d$, the candidate pool will be $X_i \cup Y_i$, where $X_i$ and $Y_i$ are irrelevant and relevant items for $i$, respectively. We format the datasets so that each instance only has one relevant item ($|Y_i| = 1$). Then, we randomly select $n$ other items to create $X_i$. The value of $n$ is determined by the test set cardinality: if it is greater than or equal to 1,000, $n$ is set to 999; otherwise, it is set to 99. The \textit{Classification} task is the only exception to this process. In this task, $|X_i| = |C_d| - 1$, where $C_d$ is the set of all possible classes for the dataset $d$ minus 1 (the correct class). In \cref{tab:dataset}, we provide details regarding the dataset cardinality, supported languages and tasks.
This benchmark creation methodology has been applied to be as faithful as possible to the one used in \textsc{MMEB}.

\section{Experiments}
\label{sec:experiments}

\begin{table*}[htb]
  \centering
  \begin{tabular}{@{}l|l|l|l|l|l|l|l@{}}
    \toprule
    \textbf{Model} & \multicolumn{7}{|c}{\textbf{Task}} \\
    \midrule
    & \textsc{I2T} & \textsc{T2I} & \textsc{VQA} & \textsc{VG} & \textsc{C} & \textsc{AVG-3} & \textsc{AVG} \\
    \midrule
    TIGER-Lab/VLM2Vec-LoRA & 13.62 & 26.21 & 49.50 & 71.40 & 8.52 & 16.12 & 33.85 \\
    \midrule
    sentence-transformers/clip-ViT-B-32-multilingual-v1 & 46.57 & 42.32 & X & X & 23.00 & 37.30 & X \\
    M-CLIP/XLM-Roberta-Large-Vit-B-16Plus & 69.88 & 67.88 & X & X & 41.00 & 59.59 & X \\
    google/siglip-base-patch16-256-multilingual & 71.10 & 68.34 & X & X & 65.87 & 68.44 & X \\
    xVLM2Vec (\textit{Ours}) & 45.02 & 48.09 & 68.00 & 73.45 & 32.02 & 41.71 & 53.32 \\
    \bottomrule
  \end{tabular}
  \caption{Average \textsc{P@1} for each task (considering all languages available for each task). \textsc{AVG} is the average \textsc{P@1} for all tasks, while \textsc{AVG-3} is the average \textsc{P@1} for the \textsc{I2T}, \textsc{T2I} and \textsc{C} tasks.}
  \label{tab:results}
\end{table*}

\begin{figure*}[ht]
      \centering
      \subfloat[Average \textsc{P@1} for \textsc{I2T}, \textsc{T2I} and \textsc{C}]{\includegraphics[width=.55\textwidth]{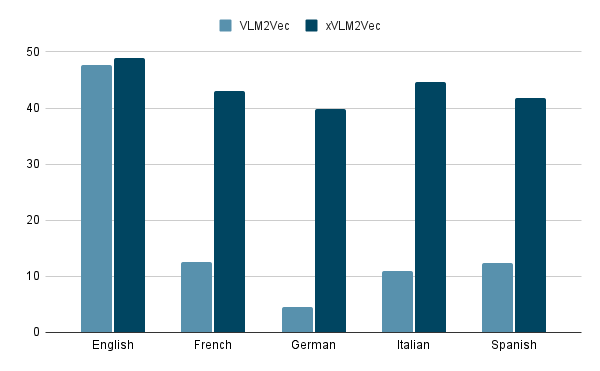}}
      \subfloat[Average \textsc{P@1} for \textsc{ALL}]{\includegraphics[width=.45\textwidth]{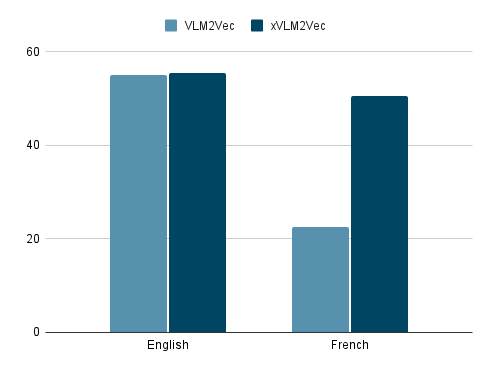}}
      \qquad
  \caption{Results for each language obtained by averaging the \textsc{P@1} on the tasks for \textsc{VLM2Vec} and \textsc{xVLM2Vec} when using the \textit{plain} formatting.}
  \label{fig:plain_results}
\end{figure*}

For our main experiments, we compare  \textsc{TIGER-Lab/VLM2Vec-LoRA} \citep{jiang2024vlm2vec}, that is the baseline English-only model checkpoint from which we start our Self-Knowledge Distillation training procedure, with our \textsc{xVLM2Vec} model. We expect this model to perform worse on non-English tasks than our trained model. 

Furthermore, we also include in the benchmark evaluation additional multilingual and multimodal models, namely:
\begin{itemize}
    \item \textsc{sentence-transformers/clip-ViT-B-32-multilingual-v1} \citep{reimers-2019-sentence-bert};
    \item \textsc{M-CLIP/XLM-Roberta-Large-Vit-B-16Plus} \citep{carlsson-EtAl:2022:LREC};
    \item \textsc{google/siglip-base-patch16-256-multilingual} \citep{zhai2023sigmoid}.
\end{itemize}

All models are either \textsc{CLIP} or \textsc{SigLIP} variants for multilingual inputs. Given the nature of these models, that is, to learn a shared embedding space that can be used to compare visual and textual embeddings, we do not test them for the \textsc{VQA} and \textsc{VG} tasks, since they require additional input modalities w.r.t. what is expected by these models (e.g. \textsc{VQA} requires both an input image and text to be compared with a collection of texts).
Furthermore, since these models are not instruction-tuned, the instances of the datasets are provided without further formatting. The only exception is the \textsc{C} task, where we add a prefix string ``image of'' (and its translation to non-English languages).

For \textsc{VLM2Vec}-based models, since they are instruction-tuned models and the embedding is extracted from the last token of the sequence (not the special eos token), we found them to be very sensitive to the formatting of the input in our initial experiments.
We will delve into this aspect in one of our ablation studies. Furthermore, while in \textsc{CLIP} and \textsc{SigLIP} models the image is represented from its visual embedding directly, in \textsc{VLM2Vec}, the image is always embedded within a string (using the special "\textless$\vert$image\_1$\vert$\textgreater" token). For our main experiments, we consider a \textit{plain} formatting by removing any punctuation at the end of the query and target strings. For the formatting text, we use the same used for the query and target in the original \textsc{MMEB} train set and their translation to the target languages that we obtain from the procedure described in \cref{sec:translation}. The formatting used for the query for each task in English is the following: 

\begin{itemize}
    \item \textsc{I2T}: \textless$\vert$image\_1$\vert$\textgreater\textbackslash n\textit{Find an image caption describing the given everyday image}
    \item \textsc{T2I}: \textit{Find me an everyday image that matches the given caption:} \{query\_text\}
    \item \textsc{VQA}: \textless$\vert$image\_1$\vert$\textgreater\textbackslash n\textit{Represent the given image with the following question:} \{query\_text\}
    \item \textsc{VG}: \textless$\vert$image\_1$\vert$\textgreater\textbackslash n\textit{Select the portion of the image that isolates the object labeled as} "\{query\_text\}"
    \item \textsc{C}: \textless$\vert$image\_1$\vert$\textgreater\textbackslash n\textit{Represent the given image for classification}
\end{itemize}

The target texts in English are formatted as follows:

\begin{itemize}
    \item \textsc{I2T}: \{target\_text\}
    \item \textsc{T2I}: \textless$\vert$image\_1$\vert$\textgreater\textbackslash n\textit{Represent the given image}
    \item \textsc{VQA}: \textit{Represent the given image with the following question:} \{target\_text\}
    \item \textsc{VG}: \textit{Select the portion of the image that isolates the object labeled as "}\{target\_text\}\textit{"}
    \item \textsc{C}: \{target\_text\}
\end{itemize}

Strings between brackets are variables that are replaced with the corresponding field in the dataset during evaluation. We provide the formatting used for non-English languages in the Appendix.

We use \textsc{P@1} as the evaluation metric, following the same setup used in the \textsc{MMEB} benchmark. We report two averages, one for the tasks that are covered by all models (\textsc{AVG-3}), that are \textsc{I2T}, \textsc{T2I} and \textsc{C}, and one for all tasks (\textsc{AVG}). We report the average results for each task type in \cref{tab:results}.

Overall, the \textsc{CLIP} and \textsc{SigLIP} models perform better than both \textsc{VLM2Vec} and \textsc{xVLM2Vec}. We believe this is due to the fact that \textsc{CLIP} and \textsc{SigLIP} models are not as general as \textsc{VLM2Vec} since the latter is an instruction-tuned model; therefore, they perform better on generic visual-text tasks. Additionally, \textsc{MMMEB} currently lacks specialized datasets, like \textsc{CIRR} \citep{liu2021image} which consists of real-life images with human-written modification sentences, due to the lack of natively multilingual and multimodal datasets. As a matter of fact, the \textsc{I2T}, \textsc{T2I} and \textsc{C} tasks all focus on generic image-text datasets (e.g. describing the contents of the image). Finally, differently from \textsc{CLIP} and \textsc{SigLIP}, \textsc{VLM2Vec} and \textsc{xVLM2Vec} can perform tasks with multiple modalities in input directly. We evaluate their performance on the \textsc{VQA} and \textsc{VG} tasks. While they do not excel in \textsc{VQA}, their performance for the \textsc{VG} task is remarkable.

\begin{table*}[htb]
  \centering
  \resizebox{\linewidth}{!}{
  \begin{tabular}{@{}l|l|l|l|l|l|l|l@{}}
    \toprule
    \textbf{Model} & \multicolumn{7}{|c}{\textbf{Task}} \\
    \midrule
    & \textsc{I2T} & \textsc{T2I} & \textsc{VQA} & \textsc{VG} & \textsc{C} & \textsc{AVG-3} & \textsc{AVG} \\
    \midrule
    TIGER-Lab/VLM2Vec-LoRA + \textit{p} & 26.89 (+13.27) & 65.06 (+38.85) & 54.82 (+5.32) & 69.21 (-2.19) & 28.10 (+19.58) & 40.02 (+23.90) & 48.82 (+14.97) \\
    xVLM2Vec (\textit{Ours}) + \textit{p} & 58.72 (+13.07) & 69.58 (+21.49) & 72.99 (+4.99) & 73.09 (-0.36) & 44.62 (+12.60) & 57.64 (+15.93) & 63.80 (+10.48) \\
    \bottomrule
  \end{tabular}}
  \caption{Average \textsc{P@1} for each task (considering all languages available for each task) for the \textsc{VLM2Vec} and \textsc{xVLM2Vec} models when using \textit{punctuation} formatting (+ \textit{p}). \textsc{AVG} is the average \textsc{P@1} for all tasks, while \textsc{AVG-3} is the average \textsc{P@1}  for the \textsc{I2T}, \textsc{T2I} and \textsc{C} tasks. We also report the increase or decrease in performance w.r.t. the corresponding model using the \textit{plain} formatting.}
  \label{tab:results_formatting}
\end{table*}

\begin{figure*}[ht]
      \centering
      \subfloat[Average \textsc{P@1} for \textsc{I2T}, \textsc{T2I} and \textsc{C}]{\includegraphics[width=.55\textwidth]{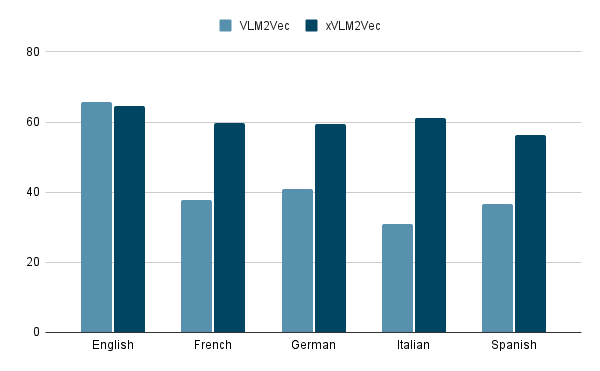}}
      \subfloat[Average \textsc{P@1} for \textsc{ALL}]{\includegraphics[width=.45\textwidth]{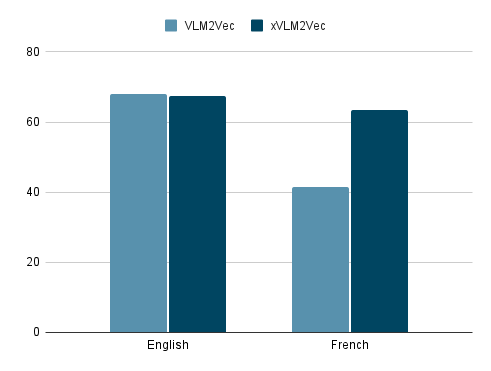}}
      \qquad
  \caption{Results for each language obtained by averaging the \textsc{P@1} on the tasks for \textsc{VLM2Vec} and \textsc{xVLM2Vec} when using \textit{punctuation} in the formatting.}
  \label{fig:punct_results}
\end{figure*}

To study the effect of our multilingual adaptation procedure, we compare the performance of the original \textsc{VLM2Vec} model and our \textsc{xVLM2Vec} model w.r.t. each language considered in the test set. Also, in this case, we distinguish between two averages. One for the \textsc{I2T}, \textsc{T2I} and \textsc{C} tasks, since they cover all the languages considered in this work, and one for all tasks, which only cover English and French. We report the results in \cref{fig:plain_results}.
Overall, we can confirm that our model \textsc{xVLM2Vec} achieves comparable results in all languages and tasks. While the \textsc{VLM2Vec} model performs well only on English language tasks. These results support our hypothesis that a language adaptation strategy improves the quality of multilingual embeddings.

\subsection{Ablation Experiments}
\label{sec:ablation}

\begin{table*}[htb]
  \centering
  \resizebox{\linewidth}{!}{
  \begin{tabular}{@{}l|l|l|l|l|l|l|l@{}}
    \toprule
    \textbf{Model} & \multicolumn{7}{|c}{\textbf{Task}} \\
    \midrule
    & \textsc{I2T} & \textsc{T2I} & \textsc{VQA} & \textsc{VG} & \textsc{C} & \textsc{AVG-3} & \textsc{AVG} \\
    \midrule
    xVLM2Vec (\textit{Ours}) + \textit{Image Loss} & 45.01 (-0.01) & 48.17 (+0.08) & 67.41 (-0.59) & 73.61 (+0.16) & 31.72 (-0.30) & 41.63 (-0.08) & 53.18 (-0.14) \\
    xVLM2Vec (\textit{Ours}) + \textit{Image Loss} + \textit{p} & 58.64 (-0.08) & 69.36 (-0.22) & 72.61 (-0.38) & 73.07 (-0.02) & 44.76 (+0.14) & 57.59 (-0.05) & 63.69 (-0.11) \\
    \bottomrule
  \end{tabular}}
  \caption{Average \textsc{P@1} for each task (considering all languages available for each task) for the \textsc{VLM2Vec} models trained with the additional image embedding loss. \textsc{AVG} is the average \textsc{P@1} for all tasks, while \textsc{AVG-3} is the average \textsc{P@1}  for the \textsc{I2T}, \textsc{T2I} and \textsc{C} tasks. We also report the increase or decrease in performance w.r.t. the corresponding \textsc{xVLM2Vec} model without image embedding loss.}
  \label{tab:results_image}
\end{table*}

We perform two ablation experiments: 1) Testing the consistency of the results w.r.t. different formatting; 2) Adding constraints to the loss at train time.

For the first ablation, since \textsc{VLM2Vec} uses the embedding of the last token, we test the effect of slight modifications to the query and target text. Specifically, while our experiments focused on a setting without punctuation in both query and target, we now add punctuation marks at the end to both of them. We add a full dot for all cases, except for the \textsc{VQA} query where we add a question mark. We will refer to this as \textit{punctuation} formatting. We report the results obtained with this setting per task in \cref{tab:results_formatting} and per language in \cref{fig:punct_results}. We find that the overall performance for both \textsc{VLM2Vec} and \textsc{xVLM2Vec} drastically increases for most tasks. Specifically, there is a significant increase in the \textsc{VQA} and \textsc{T2I} tasks w.r.t. models evaluated using the \textit{plain} formatting. Furthermore, while the performance of \textsc{VLM2Vec} increases in non-English tasks, it still falls behind w.r.t. \textsc{xVLM2Vec}. Finally, \textsc{xVLM2Vec} achieves competitive performance w.r.t. \textsc{CLIP} and \textsc{SigLIP} models in the \textsc{T2I} and \textsc{C} tasks (that is, it performs better than at least two models).

For the second ablation, we train a new model by adding constraints during training. Specifically, due to the training process focusing on the last token, the image embeddings learned by the model may be modified during training. Intuitively, we do not want the original image representation to be modified since it is language-independent. Following this intuition, we modify the original loss in \cref{eq:loss_last,eq:loss_weighted} as follows:

\begin{equation}
  loss_i(x, y) = loss_d(pool(T_i(x)), pool(S_i(x))) + loss_d(pool(T_i(x)), pool(S_i(y)))
  \label{eq:loss_image}
\end{equation}

\begin{equation}
    loss(x, y) = \frac{loss_e(x, y) + loss_i(x, y)}{4}
  \label{eq:loss_image_weighted}
\end{equation}

\noindent where $T_i(x)$ and $S_i(x)$ are the image embeddings extracted from the teacher and student model, respectively. We perform mean pooling of the image embeddings to obtain a single dense vector. The objective of $loss_i$ is to minimize the distance for the image embeddings of the student $S$ for both English and non-English inputs w.r.t. the original embeddings learned by $T$ for the English input.
We keep the Mean Squared Error for $loss_d$. 
We then compute $loss$ as the average of the four addendum (both the original $loss_e$ and $loss_i$).
We expect an improvement in performance due to the potential improvement in embedding alignment for the student model. 
We report results for the \textit{plain} and \textit{punctuation} formatting in \cref{tab:results_image}. Overall, the model performance is the same as the model trained without using $loss_i$.
To test if these performance differences are statistically significant, we use \textit{McNemar's test} \citep{McNemar1947}.
Following best practices, we use a chi-squared test with continuity correction when the sum of the cases where the output of the two models is different is more than or equal to $25$ and an exact binomial test otherwise. We perform this test for each task and for each dataset that we considered in our benchmark. 
Specifically, we compare the model trained with $loss_i$ with the one trained without $loss_i$ using both \textit{plain} and \textit{punctuation} formatting.
Overall, out of the 58 possible combinations (per task, dataset and language), we find statistical significance in the difference of the results for only 5 tasks. Therefore, adding this loss to the original methodology did not impact the training process.

\section{Conclusions}
\label{sec:conclusions}

In this work, we introduced a strategy based on a Self-Knowledge Distillation approach to make monolingual \textsc{LVLM}-based embedding models multilingual, following proper best practices introduced by related works in the textual domain. We performed experiments using \textsc{VLM2Vec}, a \textsc{Phi-3.5-Vision-Instruct} model trained for embedding extraction using contrastive learning over the embedding representation of the last token.
Results show that our approach effectively improves the model's performance in the selected non-English languages (French, German, Italian, and Spanish) while maintaining the original model's performance in English.
Furthermore, given the lack of rigorous evaluation benchmarks for multilingual and multimodal embedding models, we introduced \textsc{MMMEB}.
We release all our resources from this study (translated data, models, benchmark and code).
In future works, we will further study multilingual adaptation and evaluation by extending the training corpus to more languages and incorporating additional datasets in \textsc{MMMEB}.

\section*{Resources}

\begin{itemize}
    \item The source code is available on GitHub\footnote{\url{https://github.com/swapUniba/xVLM2Vec}}
    \item Models trained in this work are available on Huggingface\footnote{\url{https://huggingface.co/collections/swap-uniba/lvlms-for-retrieval-67d092cf93263f2062b4f16c}}
    \item Train dataset translated to French, German, Italian and Spanish is available on Huggingface\footnote{\url{https://huggingface.co/datasets/swap-uniba/xMMEB-train}}
    \item Benchmark dataset formatted for evaluation is available on Huggingface\footnote{\url{https://huggingface.co/datasets/swap-uniba/MMMEB-Benchmark}}
\end{itemize}

\section*{Acknowledgments}

We acknowledge the support of the PNRR project FAIR - Future AI Research (PE00000013), Spoke 6 - Symbiotic AI (CUP H97G22000210007) under the NRRP MUR program funded by the NextGenerationEU. 
Models are trained on the Leonardo supercomputer with the support of CINECA-Italian Super Computing Resource Allocation, class C project IscrC\_LLMM (HP10CLKWTP).

\section*{Appendix}
\label{sec:appendix}

\subsection*{Formatting}
\label{src:formatting}

In this section, we report the query and target formatting that has been used for non-English languages.

\subsubsection*{French Query}

\begin{itemize}
    \item \textsc{I2T}: \textless$\vert$image\_1$\vert$\textgreater\textbackslash n\textit{Trouvez une légende décrivant l'image donnée}
    \item \textsc{T2I}: \textit{Trouvez-moi une image de tous les jours qui correspond à la légende donnée:} \{query\_text\}
    \item \textsc{VQA}: \textless$\vert$image\_1$\vert$\textgreater\textbackslash n\textit{Représentez l'image donnée avec la question suivante:} \{query\_text\}
    \item \textsc{VG}: \textless$\vert$image\_1$\vert$\textgreater\textbackslash n\textit{Sélectionnez la partie de l'image qui isole l'objet étiqueté comme} "\{query\_text\}"
    \item \textsc{C}: \textless$\vert$image\_1$\vert$\textgreater\textbackslash n\textit{Représentez l'image donnée pour la classification}
\end{itemize}

\subsubsection*{French Target}

\begin{itemize}
    \item \textsc{I2T}: \{target\_text\}
    \item \textsc{T2I}: \textless$\vert$image\_1$\vert$\textgreater\textbackslash n\textit{Représentez l'image donnée}
    \item \textsc{VQA}: \{target\_text\}
    \item \textsc{VG}: \textless$\vert$image\_1$\vert$\textgreater\textbackslash n\textit{Représentez l'image recadrée donnée de l'objet}
    \item \textsc{C}: \{target\_text\}
\end{itemize}

\subsubsection*{German Query}

\begin{itemize}
    \item \textsc{I2T}: \textless$\vert$image\_1$\vert$\textgreater\textbackslash n\textit{Finde eine Bildunterschrift, die das gegebene Alltagsbild beschreibt}
    \item \textsc{T2I}: \textit{Finde mir ein alltägliches Bild, das der gegebenen Beschriftung entspricht:} \{query\_text\}
    \item \textsc{VQA}: X
    \item \textsc{VG}: X
    \item \textsc{C}: \textless$\vert$image\_1$\vert$\textgreater\textbackslash n\textit{Stellen Sie das gegebene Bild für die Klassifizierung dar}
\end{itemize}

\subsubsection*{German Target}

\begin{itemize}
    \item \textsc{I2T}: \{target\_text\}
    \item \textsc{T2I}: \textless$\vert$image\_1$\vert$\textgreater\textbackslash n\textit{Stelle das gegebene Bild dar}
    \item \textsc{VQA}: X
    \item \textsc{VG}: X
    \item \textsc{C}: \{target\_text\}
\end{itemize}

\subsubsection*{Italian Query}

\begin{itemize}
    \item \textsc{I2T}: \textless$\vert$image\_1$\vert$\textgreater\textbackslash n\textit{Trova una didascalia che descriva l'immagine di tutti i giorni}
    \item \textsc{T2I}: \textit{Trovami un'immagine di tutti i giorni che corrisponda alla didascalia data:} \{query\_text\}
    \item \textsc{VQA}: X
    \item \textsc{VG}: X
    \item \textsc{C}: \textless$\vert$image\_1$\vert$\textgreater\textbackslash n\textit{Rappresenta l'immagine data per la classificazione}
\end{itemize}

\subsubsection*{Italian Target}

\begin{itemize}
    \item \textsc{I2T}: \{target\_text\}
    \item \textsc{T2I}: \textless$\vert$image\_1$\vert$\textgreater\textbackslash n\textit{Rappresenta l'immagine data}
    \item \textsc{VQA}: X
    \item \textsc{VG}: X
    \item \textsc{C}: \{target\_text\}
\end{itemize}

\subsubsection*{Spanish Query}

\begin{itemize}
    \item \textsc{I2T}: \textless$\vert$image\_1$\vert$\textgreater\textbackslash n\textit{Encuentra una leyenda que describa la imagen cotidiana dada}
    \item \textsc{T2I}: \textit{Encuentra una imagen cotidiana que coincida con la leyenda dada:} \{query\_text\}
    \item \textsc{VQA}: X
    \item \textsc{VG}: X
    \item \textsc{C}: \textless$\vert$image\_1$\vert$\textgreater\textbackslash n\textit{Representa la imagen dada para clasificación}
\end{itemize}

\subsubsection*{Spanish Target}

\begin{itemize}
    \item \textsc{I2T}: \{target\_text\}
    \item \textsc{T2I}: \textless$\vert$image\_1$\vert$\textgreater\textbackslash n\textit{Representa la imagen dada}
    \item \textsc{VQA}: X
    \item \textsc{VG}: X
    \item \textsc{C}: \{target\_text\}
\end{itemize}

\bibliographystyle{unsrtnat}
\bibliography{references}  






\end{document}